\documentclass[letterpaper,10pt,conference]{ieeeconf}

\IEEEoverridecommandlockouts
\usepackage[T1]{fontenc}
\usepackage{graphicx}
\usepackage{amsmath}
\usepackage{amssymb}
\usepackage{array}
\usepackage{url}
\usepackage{booktabs}
\usepackage{makecell}
\usepackage{cite}
\usepackage{flushend}
\usepackage[protrusion=true,expansion=false]{microtype}
\usepackage[htt]{hyphenat}
\usepackage{wrapfig}
\usepackage{subcaption}
\let\labelindent\relax
\usepackage{enumitem}

\graphicspath{{figures/}}

\newcommand{\mps}{\,\mathrm{m/s}}
\newcommand{\code}[1]{\texttt{#1}}

\title{\LARGE \bf
Reference-Free Humanoid Skateboarding via Termination-Shaped Reinforcement Learning
}

\author{Ganeshvar Soenessardien, Michael Przystupa and Kevin Sebastian Luck%
\thanks{All authors are with the Department of Artificial Intelligence, Faculty of Science, Vrije Universiteit Amsterdam, De Boelelaan 1105, 1081~HV Amsterdam, The Netherlands. {\tt\small g.k.p.soenessardien@student.vu.nl}, {\tt\small m.v.przystupa@vu.nl}, {\tt\small k.s.luck@vu.nl}}%
}

\begin{document}

\maketitle
\thispagestyle{empty}
\pagestyle{empty}

\begin{abstract}
Humanoid locomotion research has advanced rapidly on walking and running, but \emph {object-mediated} locomotion, in which a robot must coordinate its body with a passive external object, remains difficult. 
One challenging task is skateboarding, which requires simultaneous interaction with the ground and a passive but movable platform to achieve forward motion. 
Current methods for tackling skateboarding with humanoid robots either require human demonstrations or utilise analytic steering models that require system identification. 
In this paper, we introduce a deep reinforcement learning framework that does not rely on prior data such as demonstrations and enables robots such as the K1 humanoid to learn to skateboard using model-free learning. 
%
The method combines task rewards with exploit-blocking terminations. 
We introduce a number of constraints restricting behaviours such as losing the planted board foot, standing with both feet on the board, excessive twisting, separation from the board, and unstable board motion. 
Their changing prevalence during training produces an implicit progression from stance control towards sustained robot--board locomotion. 
Experiments are performed on a simulated K1 humanoid robot in Mujoco. 
The results demonstrate the feasibility of reference-free humanoid-object locomotion with modest hardware constraints while identifying limitations in movement style, statistical precision, reproducibility and eventual hardware transfer.
\end{abstract}

\section{Introduction}
\label{sec:intro}


Humanoid robots can now walk, balance, recover from pushes and traverse rough terrain using learned controllers trained in massively parallel simulators \cite{doi:10.1126/scirobotics.adi9579,seo2025learningsimtorealhumanoidlocomotion,cui2024adapting,gu2024advancinghumanoidlocomotionmastering}. Almost all of these skills involve \emph{direct} contact between the robot's feet and a static support surface. 
However, a large and practically important class of behaviours is more difficult to master: locomotion tasks in which a robot has to interact with a passive object, that is neither actuated nor rigidly attached to the body, to achieve movement. 
Pushing a cart, riding a scooter, balancing on a mobility aid, or skateboarding all share this structure. 
The object stores momentum, introduces its own contact dynamics and can fail in ways the robot does not always directly control \cite{humanoidlocomanipulationspattern2021,hawley2017controlstrategy,rioux2015humanoid,BALTES2023106941,han2026huskyhumanoidskateboardingphysicsaware}.

Skateboarding is a particularly demanding instance of object-mediated locomotion and therefore a useful benchmark for embodied artificial intelligence. 
To locomote with a skateboard, a humanoid must keep one foot on a small, freely rolling deck, and use the other foot to generate propulsion against the ground while regulating the contacts among three coupled bodies (robot, board and ground). 
Misplaced contacts can induce foot slip, unwanted board rotation, incorrect foot-board contact patterns, failed phase transitions, or degeneration into merely pushing the board rather than riding it. 
Success therefore requires contact rich whole-body control, robot-object coordination and stable locomotion through an under-actuated external system simultaneously \cite{han2026huskyhumanoidskateboardingphysicsaware,thibault2024learningskateboardinghumanoidrobots,takasugi2016realtimeskatingmotion,liu2025discretetimehybridautomatalearning}.
\begin{figure}
    \centering
    \includegraphics[width=0.5\linewidth]{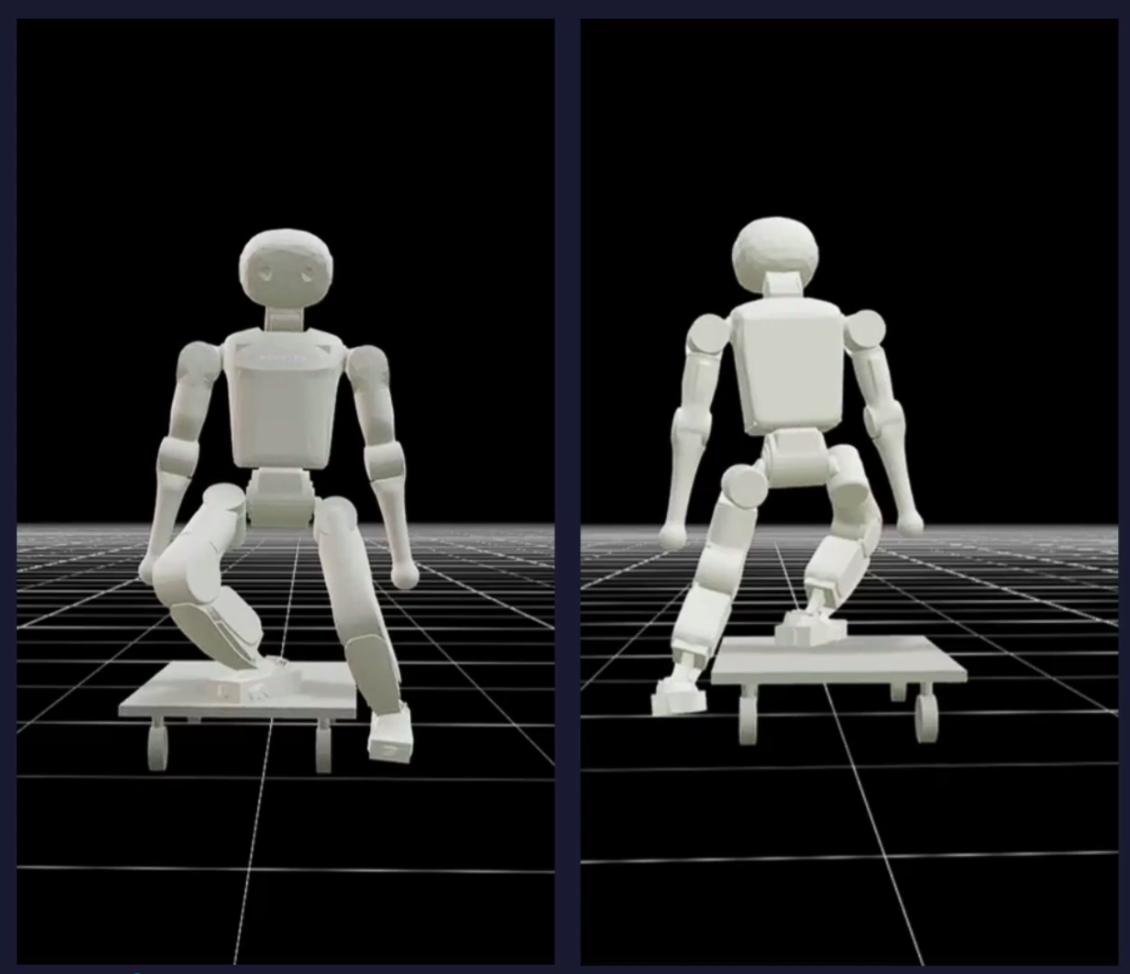}
    \caption{This paper investigates the problem of learning to skateboard on mobile platforms (above: front- and back-view) in humanoids (K1 Booster) with model-free reinforcement learning. }
    \label{fig:placeholder}
\end{figure}
%
This paper investigates applying reinforcement learning to humanoid locomotion on a skateboard-like platform without the use of task-specific priors used by recent humanoid skateboarding systems. 
HUSKY, for example, achieves agile real-world humanoid skateboarding by combining Adversarial Motion Priors (AMP) \cite{peng2021amp} from human pushing demonstrations with a physics-guided lean-to-steer model and trajectory-guided phase transitions, and validates the resulting controller on real hardware \cite{han2026huskyhumanoidskateboardingphysicsaware}. 
Such reference data and engineering are powerful when available, but they also make the method depend on human demonstrations, physical modelling and task-specific integration which are often not available or hard to obtain. 
Furthermore, such data may be hard to utilize due to the existence of the correspondence problem between robot and human embodiment \cite{aktas2024correspondence, arduengo2021human}. 

This paper focuses on the harder and complementary case of reference-free learning \cite{siekmann2021simtoreallearningcommonbipedal}. The controller receives no motion-capture trajectory, no target joint-posture sequence, and no explicit posture-tracking reward that tells it what skating should look like. 
The skateboarding behaviour must be selected and emerge by learning from the task reward, the reset geometry and the early termination conditions. 
The question is whether recognisable humanoid-board locomotion can emerge from this minimal supervision alone and what behaviour appears when no human motion prior is present to shape its style.
\newline\textbf{Contribution: } %
The objective is to design, train, and evaluate a reference free reinforcement-learning controller that enables a Booster K1 humanoid to propel a passive four-wheel roller board forward while remaining upright, keeping one foot on the board and using the other to push on the ground. Three research questions structure the work:

\begin{enumerate}[label=\textbf{RQ\arabic*}, ref=RQ\arabic*, leftmargin=*, wide=0pt]
  \item \emph{Feasibility and extent.} To what extent can a Booster K1 humanoid learn
        recognisable skateboard propulsion without any reference motion or explicit
        posture-tracking reward, and what is the character of the resulting behaviour?
        \label{rq:feasibility}
  \item \emph{Mechanism.} Which reward and termination design choices are necessary to
        prevent trivial exploit solutions and steer the policy toward skating-like
        mechanics? \label{rq:mechanism}
  \item \emph{Consistency and cost.} How consistent is the learned behaviour across
        independent seeds when trained under a constrained hardware budget, and what does
        that budget permit and preclude? \label{rq:consistency}
\end{enumerate}
\section{Related Work}
\label{sec:related}

\subsection{Humanoid and Legged Locomotion}

The current generation of learned legged controllers is enabled by GPU-accelerated simulators that run thousands of environments in parallel, including Isaac Gym \cite{makoviychuk2021isaacgymhighperformance} and MuJoCo-based frameworks such as mjlab, which is used in this paper \cite{mujoco,zakka2026mjlablightweightframeworkgpuaccelerated}. Rudin et al. \cite{rudin2022learningwalkminutesusing} learned quadruped walking in minutes with massively parallel PPO, and learned humanoid walking controllers have since been transferred to real hardware \cite{singh2023currentfeedback,doi:10.1126/scirobotics.adi9579}. These works establish the simulation-and-PPO recipe adopted here, but target locomotion over \emph{static terrain}; this paper adds a passive, freely rolling object into the support chain.

\subsection{Reference-Based and Reference-Free Motion Learning}

A common strategy for natural looking physics-based motion is to supervise the policy with human or animal motion data. DeepMimic tracks reference clips using an imitation reward \cite{Peng_2018}, while Adversarial Motion Priors (AMP) replace explicit trajectory tracking with a learned discriminator that rewards motions matching the statistics of a reference dataset \cite{peng2021amp}. These methods are powerful because they inject a strong prior over \emph{what motion should look like}. The complementary line, to which this paper belongs, removes that prior: Heess et al. showed that rich locomotion behaviours can \emph{emerge} from task reward and environment design alone \cite{heess2017emergence}. This reference-free setting is often harder because the reward can be underspecified and may admit many qualitatively different solutions \cite{siekmann2021simtoreallearningcommonbipedal}. In particular, without a reference it is difficult to converge to smooth motion that does not exploit simulator artefacts. Characterising the solution that emerges for humanoid skateboarding is a central aim here.

\subsection{Robots on Dynamic Platforms and Skateboarding}

Skating and rolling-platform locomotion have a long history in model-based robotics. Classical work analyses the skateboard as a human-controlled non-holonomic system \cite{HUBBARD1980745,varszegi2015skateboard}, while humanoid skating has been controlled using model-based real-time balance and foot-force control \cite{takasugi2016realtimeskatingmotion}. More recently, learning-based and optimisation-based methods have been applied to legged robots on dynamic platforms: quadrupeds have been taught to skateboard through optimisation-based control \cite{xu2024optimization} and to mount a skateboard with reverse-curriculum learning \cite{belov2025quadrupedal}; hybrid automata learning has been applied to legged skateboarding \cite{liu2025discretetimehybridautomatalearning} and roller-skate-style swizzle gaits have been synthesised for a humanoid \cite{gu2026skater}.

The most direct point of comparison is HUSKY, a physics-aware whole-body controller that enables a Unitree G1 humanoid to perform agile, humanlike skateboarding with pushing, lean-to-steer turning and phase transitions, validated in both simulation and the real world. The HUSKY system is an instructive contrasting system to the work presented in this paper. 
HUSKY achieves a natural stroke by learning an AMP style reward from human pushing demonstrations \cite{peng2021amp}, derives steering from an explicit board-tilt/truck-steering kinematic model, plans trajectories to bridge phases, and performs sim-to-real identification of the physical board \cite{han2026huskyhumanoidskateboardingphysicsaware}. This paper instead removes the motion prior entirely, restricts the task to a single push phase, replaces analytic steering and trajectory planning with a termination-shaped feasibility set, and stays in simulation on consumer hardware. The reference-free result here is therefore best read as a lower bound on what minimal supervision can achieve and provides an empirical explanation on the value motion priors provide for producing natural, agile skating.

\section{Problem Statement}
We investigate the problem of learning how to locomote forward using a passively moved mobile platform (i.e., skateboarding) with reinforcement learning. We model the problem as a contextual Markov Decision Process (cMDP) $(\mathcal{S},\mathcal{A},\mathcal{C},p(\mathbf{c}),p(s_0), p(s_{t+1}\vert s_t,a_t), r(s_t,a_t,\mathbf{c}),\gamma)$ \cite{hallak2015contextual,benjamins2023contextualize}.
The physical state is represented by 44 generalized positions and 42
generalized velocities, hence $\mathcal{S}\subset\mathbb{R}^{86}$.
Our context
$\mathbf{c}=(v_x^*,v_y^*,\omega_z^*)\in
\mathcal{C}\subset\mathbb{R}^{3}$
is the desired velocity command. In the experiments,
$v_x^*$ is sampled uniformly from $[0.2,0.6]\,\mathrm{m/s}$ along the board's body $x$ axis, while $v_y^*=\omega_z^*=0$.
The policy acts on an observation $o_t$ and $\mathbf{c}$. Our objective is
to optimize the expected return over the context distribution,
\begin{equation}
\label{eq:objective}
\underset{\mathbf{c}\sim p(\mathbf{c}), \tau\sim p(\tau\vert\mathbf{c})}{\mathbb{E}}
\left[\sum_{t=0}^{T}\gamma^{t} r(s_t,a_t,\mathbf{c})\right],
\end{equation}
where
$p(\tau\vert\mathbf{c}) = p(s_0)\prod_{t=0}^{T-1}
\pi_\theta(a_t\vert o_t,\mathbf{c})\,p(s_{t+1}\vert s_t,a_t)$
is the trajectory distribution induced by the policy under context $\mathbf{c}$.

\subsection{Robot, Board, and World}

The task 
is implemented in mjlab \cite{zakka2026mjlablightweightframeworkgpuaccelerated} using MuJoCo physics \cite{mujoco} and NVIDIA Warp acceleration \cite{nvidiawarpdocs}. 
The robot is the Booster K1 humanoid \cite{boosterk1}. 
The environment is a flat $4\,\mathrm{m}\times4\,\mathrm{m}$ plane containing an unactuated four-wheel TARROX Atlas transport roller board ($590\times390\times120\,\mathrm{mm}$ overall, with an $18\,\mathrm{mm}$ deck thickness and a mass of $4.22\,\mathrm{kg}$) \cite{hornbachtarroxatlas}, which the robot can influence only through foot contact. 
Episodes are capped at $1{,}000$ control steps ($20\,\mathrm{s}$ at a $50\,\mathrm{Hz}$ policy rate over $200\,\mathrm{Hz}$ physics).

\subsection{Initial State: the Push Stance}
\label{sec:initstate}

\begin{wrapfigure}{l}{0.2\textwidth}
  \begin{center}
    \includegraphics[width=0.2\textwidth]{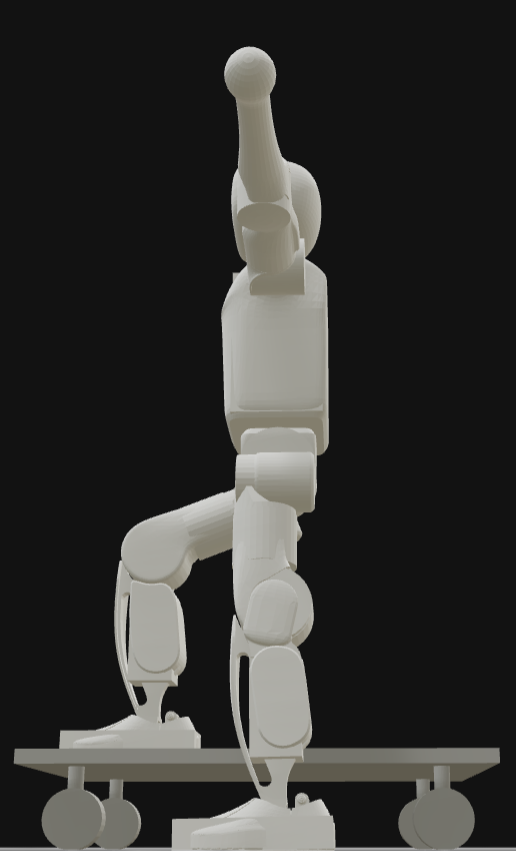}
  \end{center}
  \caption{Initial stance of the robot at $s_0$.}
  \label{fig::startstate}
\end{wrapfigure}
Each episode begins in a push stance with one foot on the board deck and the other beside the board on the ground (Fig. \ref{fig::startstate}). The robot trunk is initialised at a height of $0.55\,\mathrm{m}$, which places the on-board foot flat on the deck surface without interpenetration. The on-board leg is flexed (hip pitch $-1.20\,\mathrm{rad}$, knee pitch $+1.20\,\mathrm{rad}$) into a hip-back, shin-vertical, flat-foot pose, while the off-board leg is nearly straight with a small outward hip roll. A lateral offset of about $0.14\,\mathrm{m}$ places the off-board foot beside the deck. The left/right foot assignment is mirrored across environments for variation and small reset jitter is added (board yaw $\pm0.03\,\mathrm{rad}$, robot yaw $\pm0.05\,\mathrm{rad}$, robot planar position $\pm0.015\,\mathrm{m}$). The board starts at rest. We call this configuration the XOR stance: one foot is inside the deck footprint, and the other is outside it and below $0.08\,\mathrm{m}$, so one foot is on the deck and the other is near the ground. The initial pose in Fig.~\ref{fig::startstate} is therefore an XOR stance, and the reward encourages the policy to maintain it (Section~\ref{sec:reward}). The XOR-stance rate is the fraction of control steps in an episode spent in this stance. 



\section{Methods}
\label{sec:method}

We solve the previously described MDP using proximal policy optimisation (PPO) \cite{schulman2017proximal} setup with two task specific components: a reward stack that favours conservative skating over standing still or kicking the board (Section~\ref{sec:reward}) and early terminations whose changing dominance during training induces an implicit curriculum (Section~\ref{sec:term}).

\subsection{Observation Design}

The actor receives an 89 dimensional observation $o_t$ concatenated with the three-dimensional command $\mathbf{c}$, giving a 92 dimensional input. Table~\ref{tab:observations} lists its components and
corruptions. The critic receives 102 inputs: noise-free observations
together with privileged foot air times, contact states and contact
forces. Terrain-scan and foot-height inputs are omitted because the
terrain is flat. The board state and contact flags in the actor input are not directly measured on hardware, so deployment would require a board-state estimator, e.g., fiducial tracking of the deck \cite{apriltag}, or providing them only to the critic, as in HUSKY \cite{han2026huskyhumanoidskateboardingphysicsaware}.

\begin{table}[!h]
\vspace{4pt}
\caption{Actor observations. Noise is independently sampled additive
uniform noise $\mathcal{U}[-b,b]$. Delays are sampled in discrete
50-Hz control steps (20\,ms per step).}
\label{tab:observations}
\centering
\scriptsize
\setlength{\tabcolsep}{2pt}
\resizebox{\columnwidth}{!}{%
\begin{tabular}{@{}p{0.46\columnwidth}p{0.33\columnwidth}c@{}}
\toprule
Observation (dimension) & Noise half-range $b$ & Delay (steps) \\
\midrule
Base linear/angular velocity (3+3)
  & 0.5\,m/s; 0.2\,rad/s & 1--2 \\
Projected gravity (3)
  & 0.05 & -- \\
Joint position/velocity (22+22)
  & 0.01\,rad; 1.5\,rad/s & -- \\
Previous action, command (22+3)
  & -- & -- \\
Board relative position (3)
  & 0.02\,m & 1--3 \\
Board planar velocity, yaw rate (2+1)
  & 0.1\,m/s; 0.1\,rad/s & 1--3 \\
Board angular velocity (3)
  & 0.1\,rad/s & 1--3 \\
Board projected gravity (3)
  & 0.03 & 1--2 \\
Board foot-contact flags (2)
  & -- & -- \\
\bottomrule
\end{tabular}}
\vspace{-15pt}
\end{table}

\subsection{Action Space and Physics}

The actor outputs $\mathbf{a}_t\in\mathcal{A}=\mathbb{R}^{22}$, with no
action clipping. The outputs are scaled into PD joint-position targets; the
ten arm and head channels have zero scale to match the walker pre-training, so only the twelve leg channels
affect the robot. Contacts are simulated using MuJoCo's soft-contact dynamics with frictional constraints \cite{mujoco}. In the MJCF model, the foot collision boxes use a sliding-friction coefficient of $0.8$, the board deck $0.6$ (density $800\,\mathrm{kg/m^3}$, contact priority $1$) and the caster wheels $0.8$ with six-dimensional contact friction. Following dynamics-randomisation approaches to sim-to-real transfer \cite{Peng_2018_sim_to_real}, start-up randomisation scales PD gains by $[0.85,1.15]$, joint damping and friction by $[0.8,1.2]$, all roller-geometry friction by $[0.7,1.3]$ and the board body mass by $[0.85,1.15]$.

\subsection{PPO Training Setup}

The actor and critic are neural networks with hidden dimensions $(512, 256, 128)$, ELU activations, and input normalisation. The actor parameterises a diagonal Gaussian policy with an observation-dependent mean and a learned, observation-independent log standard deviation for each of the 22 action dimensions, initialised to $0$ (standard deviation $1.0$); exponentiation ensures positivity.
The critic produces a deterministic estimate of the value. PPO uses $24$ rollout steps per environment, $5$ learning epochs, $8$ minibatches, discount $\gamma=0.99$, GAE $\lambda=0.95$, entropy coefficient $0.005$, clip parameter $\epsilon=0.2$, clipped value loss with coefficient $1.0$, maximum gradient norm $1.0$ and an adaptive learning rate initialised at $10^{-3}$ with a target KL divergence of $0.01$. Each run uses $6{,}144$ parallel environments. The actor is initialised from the generic walker policy; the critic and optimiser start fresh, and no run is warm-started from a prior skateboarding policy.

\subsection{Reward Design}
\label{sec:reward}

The reward stack (Table~\ref{tab:reward}) favours forward board--body motion in a stable push stance. With command direction $\hat{\mathbf{c}}$, board velocity $\mathbf{v}_{\mathrm{board}}$, trunk height $z_{\mathrm{trunk}}$, control timestep $\Delta t$ and indicator function $\mathbf{1}[\cdot]$, the main term is
\begin{equation}
\label{eq:board}
r_{\mathrm{board}}=40\Delta t\,\max(0,\mathbf{v}_{\mathrm{board}}\!\cdot\!\hat{\mathbf{c}})\mathbf{1}[z_{\mathrm{trunk}}>0.50\,\mathrm{m}].
\end{equation}
The term is zero when the trunk drops below $0.50\,\mathrm{m}$. A companion term rewards forward trunk motion only in the XOR stance (Section~\ref{sec:initstate}), with weight $15$. Upright, trunk height and XOR-bonus terms maintain this stance.

The near-ground term rewards the lowest off-board foot:
\begin{equation}
\label{eq:nearground}
r_{\mathrm{ground}}=4\Delta t\,\exp\!\left[-\frac{\min_{i\in\mathcal{F}_{\mathrm{off}}}z_{\mathrm{foot},i}^{2}}{(0.20\,\mathrm{m})^2}\right].
\end{equation}
The reward is zero if $\mathcal{F}_{\mathrm{off}}$ is empty. It was introduced to support early stability but also favours the low clearance shuffle analysed in Section~\ref{sec:why}.

\begin{table}[t]
\vspace{0.2cm}
\caption{Reward terms used for skateboarding.}
\label{tab:reward}
\centering
\scriptsize
\begin{tabular*}{\columnwidth}{@{}l@{\extracolsep{\fill}}rl@{}}
\toprule
Term & Weight & Purpose \\
\midrule
Upright & $+12$ & Trunk verticality \\
Trunk height & $+15$ & Reward height near $0.55\,\mathrm{m}$ \\
Board distance & $+40$ & Forward board motion \\
Body progress & $+15$ & Forward trunk motion in XOR \\
XOR stance & $+5$ & Maintain push stance \\
Near ground & $+4$ & Keep off-board foot low \\
Idle on board & $-10$ & Block stationary deck standing \\
Wrong way & $-4$ & Penalise backward motion \\
Joint limits & $-1$ & Enforce soft limits \\
Action rate & $-0.01$ & Smooth actions \\
Self-collision & $-0.1$ & Discourage self-contact \\
\bottomrule
\end{tabular*}
\vspace{-0.1cm}
\end{table}

\subsection{Termination-Shaped Curriculum}
\label{sec:term}

Unlike the standard fall and time-limit terminations, the task-specific terminations below restrict which robot--board interactions remain feasible. Such terminations act as hard constraints; constrained RL formalises them explicitly~\cite{zhang2022penalizedproximalpolicyoptimization,sun2022neuripsw-constrained,kim2024rewardsconstraintsapplicationslegged}, and CaT~\cite{chanesane2024catconstraintsterminationslegged} likewise enforces locomotion constraints through terminations. We use plain terminations without a constraint formalism and leave such a formulation to future work. The principal curriculum driver is \code{board\_foot\_displaced}. At reset, the initial board-frame position of the on-board foot is recorded and the episode terminates if it moves by more than $5$~cm. This condition dominates early training and pressures the policy to plant its board foot before collecting sustained board-motion reward. As stance control improves and the board travels farther, \code{board\_out\_of\_bounds} becomes more prominent. This shift in the dominant failure mode creates an implicit curriculum from stance control to locomotion without changing the environment during training.

The remaining terminations remove specific failure modes (Table~\ref{tab:term}). \code{both\_feet\_on\_board} blocks standing on the deck; \code{robot\_board\_yaw\_misaligned} blocks excessive body--board twisting; and \code{trunk\_too\_low} blocks crouched and split-crawl solutions. Board inversion, excessive trunk rotation and robot--board separation are handled by \code{board\_inverted}, \code{spinning\_out} and \code{robot\_far\_from\_board}, respectively. The expected learning pattern is therefore that board-foot displacement dominates initially, XOR stance rises, board and trunk speeds increase together, and time outs become more frequent as episodes lengthen.

Development observations suggest that \code{board\_foot\_displaced} and \code{robot\_board\_yaw\_misaligned} were particularly important for producing a planted board foot and maintaining body--board alignment. However, the terminations were not individually ablated, so their marginal contributions cannot be established experimentally and remain future work.

\begin{table}[t]
\vspace{0.2cm}
\caption{Early terminations. Their changing prevalence produces the implicit curriculum.}
\label{tab:term}
\centering
\scriptsize
\setlength{\tabcolsep}{4pt}
\renewcommand{\arraystretch}{1.02}
\resizebox{\columnwidth}{!}{%
\begin{tabular}{@{}lll@{}}
\toprule
Termination & Threshold & Role \\
\midrule
\code{board\_foot\_displaced} & displacement $>5$~cm & Keep board foot planted \\
\code{board\_out\_of\_bounds} & distance $>1.4$~m & Bound forward travel \\
\code{both\_feet\_on\_board} & both feet on deck & Block standing on board \\
\code{robot\_board\_yaw\_misaligned} & yaw error $>0.4$~rad & Block twisting \\
\code{trunk\_too\_low} & trunk $z<0.40$~m & Block collapsed postures \\
\code{robot\_far\_from\_board} & distance $>0.7$~m & Block separation \\
\code{board\_inverted} & tilt $>60^\circ$ & Block board flipping \\
\code{spinning\_out} & ang.\ speed $>8$~rad/s & Block spinning \\
\code{fell\_over} & tilt $>50^\circ$ & Block severe body tilt \\
\code{nan\_detection} & non-finite state & Numerical safety \\
\code{time\_out} & $1{,}000$ steps ($20$~s) & Cap episode duration \\
\bottomrule
\end{tabular}}
\vspace{-0.2cm}
\end{table}

\section{Experimental Design}
\label{sec:experiments}

\subsection{Training Protocol and Seeds}

Experiments started with a pre-trained walking policy as warm-start, which defines a standard walking motion but was learned \textbf{without} the moving platform being present. The walker is never used as a tracking target but still introduces a walking style prior; whether this warm start is necessary was not tested. 
All experiments used identical high-level settings, $6{,}144$ parallel environments and a target of $20{,}000$ PPO iterations and were repeated three times. 
Each run completed approximately $19{,}800$--$20{,}000$ PPO iterations.
   

\subsection{Hardware and Compute Budget}

We used a single workstation (NVIDIA RTX~5070~Ti GPU, AMD Ryzen~9~7900 with 12 cores/24 threads, 32~GB RAM). At $6{,}144$ environments, each iteration took $\approx\!5.1$--$5.3$~s once warm, so one $20{,}000$-iteration run took $\approx\!28.7$~h. 

\subsection{Metrics}

Training progress is measured by average episode reward and episode length. Skating mechanics are evaluated using episode averages of four per-step metrics: board forward speed, grounded body progress (forward trunk velocity in XOR stance), trunk forward velocity and XOR-stance rate. Failure modes are tracked through termination counts for board out of bounds, board-foot displacement, yaw misalignment, both feet on the board and time out. Curves are smoothed only for visualisation; the exported CSV files retain the raw logged values and final statistics are computed from the unsmoothed final value of each seed.





\section{Results}
\label{sec:results}

\subsection{Learning Progress}

\begin{figure}[!t]
\vspace{0.2cm}
\centering
\begin{subfigure}[b]{0.32\textwidth}
  \centering
  \includegraphics[width=\textwidth]{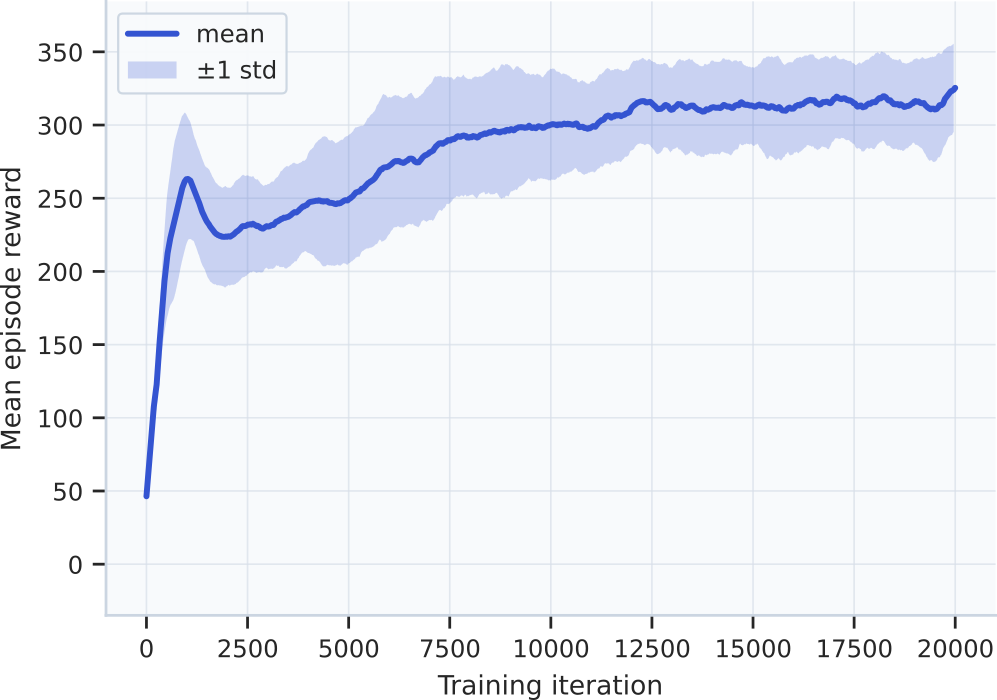}
  \caption{Episodic reward}
  \label{fig:episodic-reward}
\end{subfigure}

\begin{subfigure}[b]{0.32\textwidth}
  \centering
  \includegraphics[width=\textwidth]{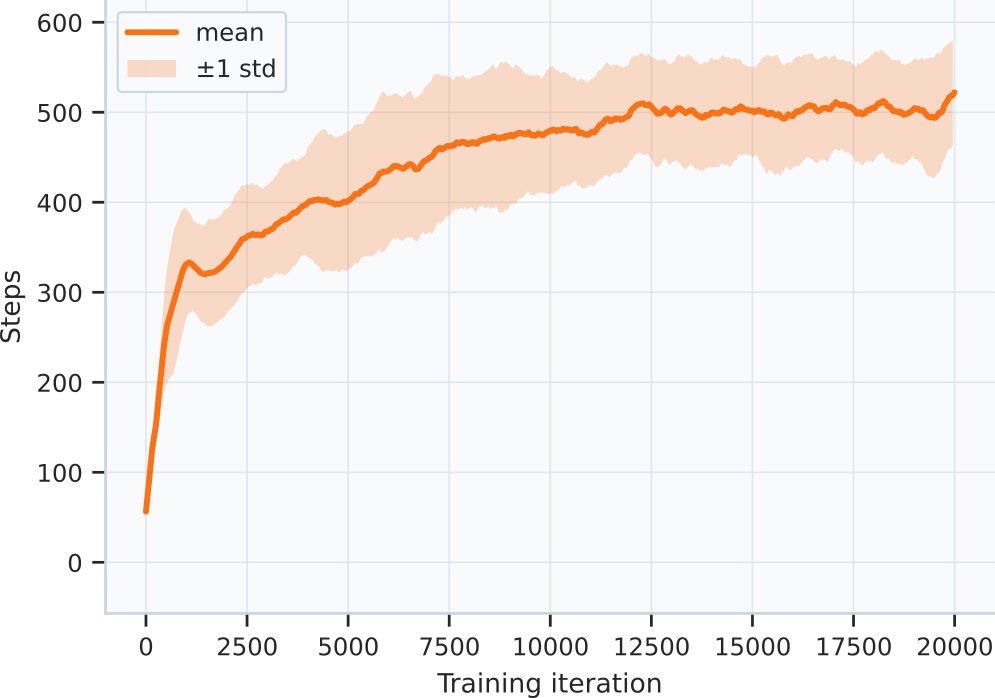}
  \caption{Steps per episode}
  \label{fig:episode-steps}
\end{subfigure}
\caption{\textbf{Learning progress} across three experiments (mean $\pm1$ sample std.). Longer episodes indicate progressively more stable rollouts.}
\label{fig:progress}
\vspace{-0.2cm}
\end{figure}

Figure~\ref{fig:progress} shows the mean and sample standard deviation of reward and episode length across the three seeds. At the final logged values, reward reaches $313.0 \pm 36.4$, while episode length reaches $498.8 \pm 76.5$ steps. Two patterns are notable. Firstly, increasing episode length indicates that the policies remain within the permitted task conditions for progressively longer periods. Secondly, both curves show an early transient: reward peaks near iteration $1{,}000$, subsequently falls and then recovers towards a plateau. As the following figures show, this transient coincides with high initial board speed and frequent board-foot displacement. The warm started policies can therefore move the board early, but sustained motion appears only once board-foot displacement decreases.


\subsection{Emergent Skating Mechanics}

\begin{figure*}
\centering
\vspace{0.6cm}
\begin{subfigure}[b]{0.24\textwidth}
         \centering
         \includegraphics[width=\textwidth]{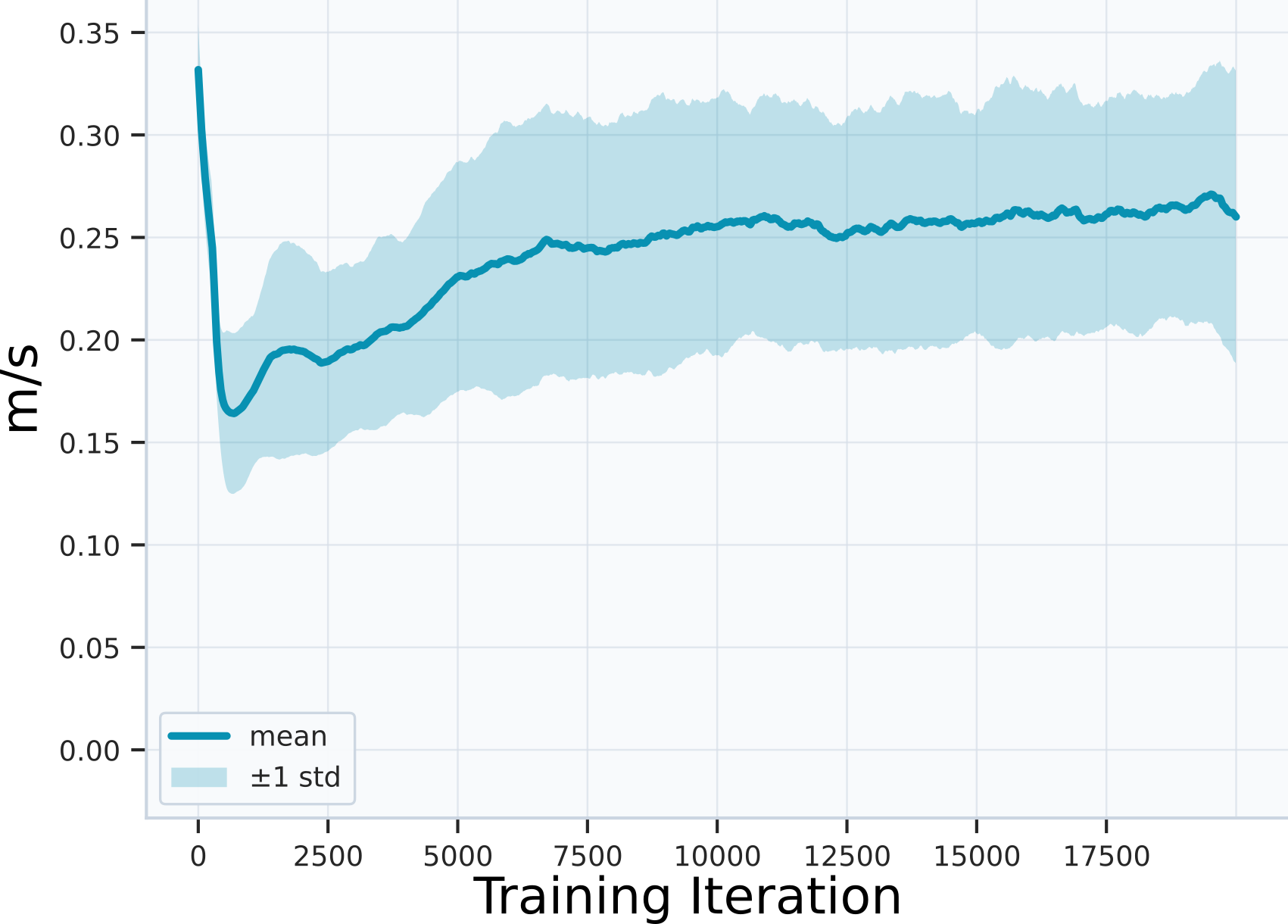}
         \caption{Board speed}
     \end{subfigure}  %
\hfill %
\begin{subfigure}[b]{0.24\textwidth}
         \centering
         \includegraphics[width=\textwidth]{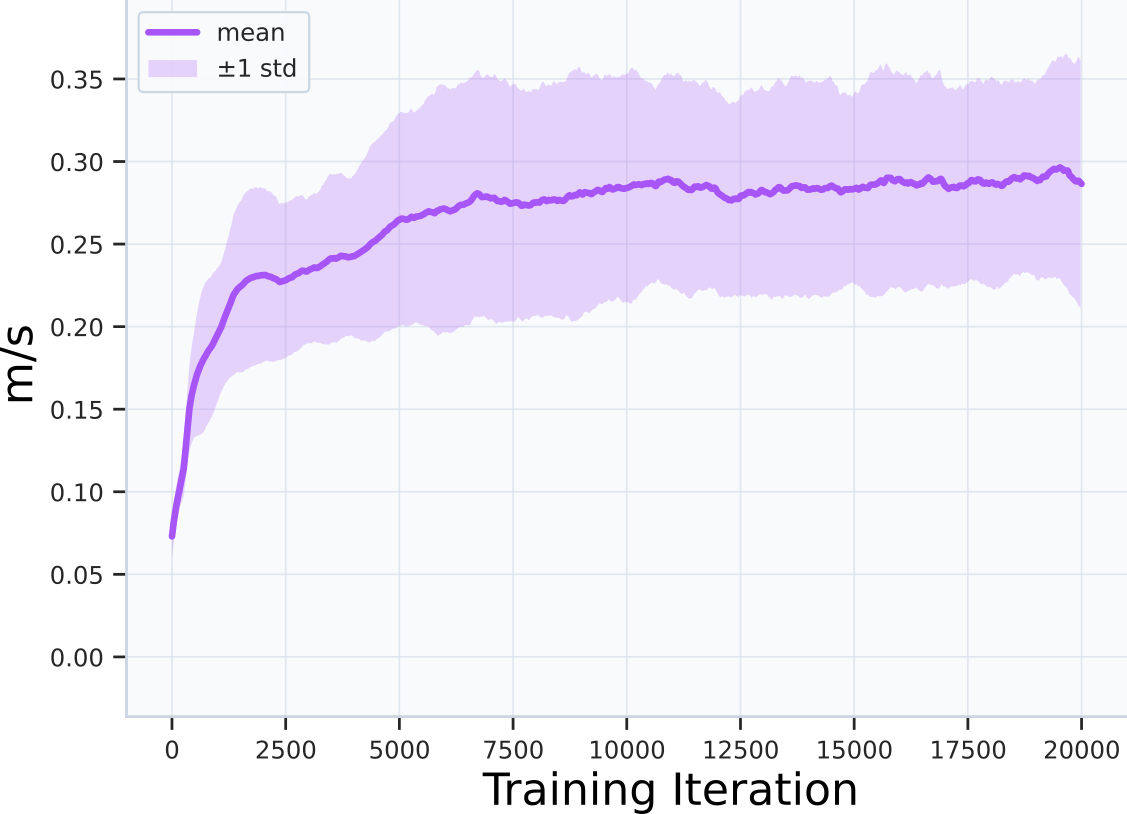}
         \caption{Push progress}
     \end{subfigure} %
\hfill %
\begin{subfigure}[b]{0.24\textwidth}
         \centering
         \includegraphics[width=\textwidth]{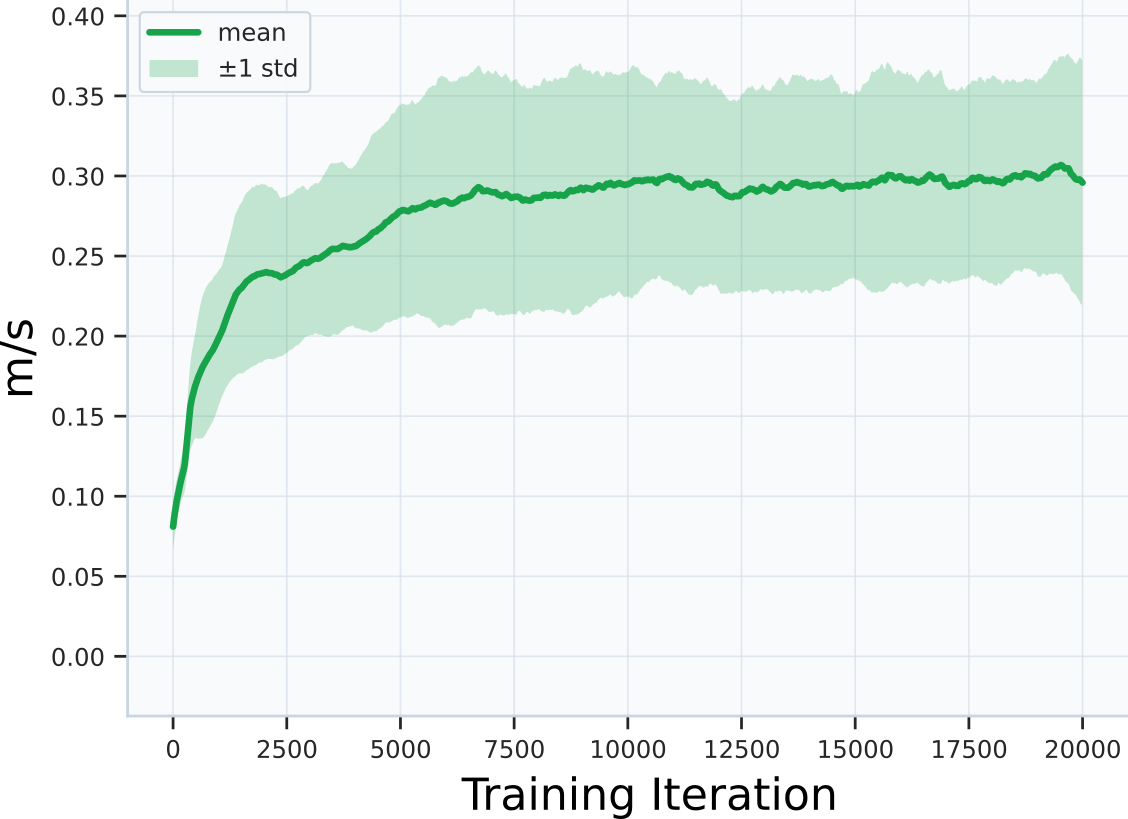}
         \caption{Trunk speed}
     \end{subfigure} %
\hfill %
\begin{subfigure}[b]{0.24\textwidth}
         \centering
         \includegraphics[width=\textwidth]{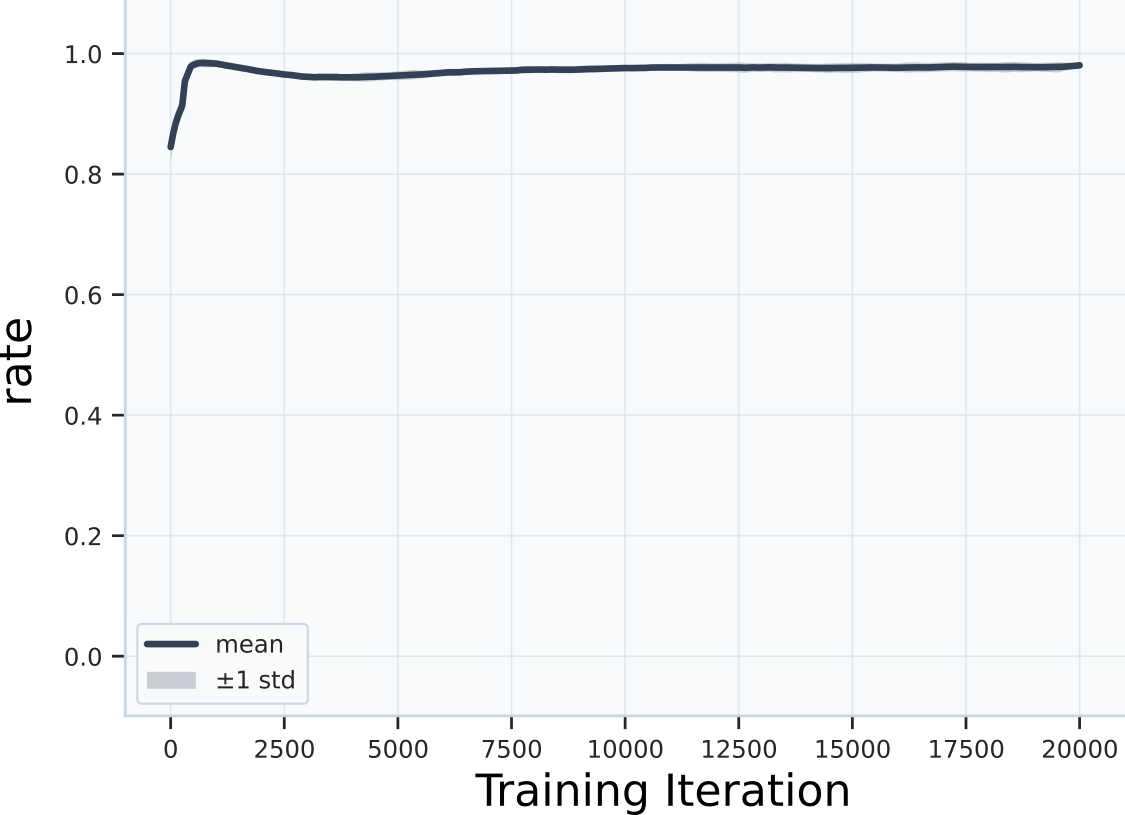}
         \caption{XOR stance}
     \end{subfigure} %
\caption{\textbf{Emergent skating mechanics across training.} Board speed, grounded body progress and trunk speed rise together while the XOR-stance rate saturates near $0.98$, indicating genuine robot--board locomotion rather than board kicking. Bands show $\pm 1$ standard deviation.}
\label{fig:mechanics}
\vspace{-0.25cm}
\end{figure*}

Figure~\ref{fig:mechanics} reports the skating-mechanics metrics. The final means are $0.270 \pm 0.087\mps$ for board speed, $0.297 \pm 0.092\mps$ for grounded body progress and $0.308 \pm 0.095\mps$ for trunk speed. Board and trunk speeds increase over the same training period and remain comparable, consistent with the robot travelling with the board, as opposed to merely launching it away. The position based XOR-stance rate exceeds $0.97$ by approximately iteration $500$ and remains near this level, reaching a final value of $0.976 \pm 0.009$. All three seeds therefore reach the intended one-foot-on-board, one-foot-near-ground stance geometry. Together with the coupled velocity metrics and qualitative inspection, this supports the interpretation that the policies learned skating like mechanics. Board speed also shows an early high speed transient, followed by a dip while stance control consolidates, consistent with Figures~\ref{fig:progress} and~\ref{fig:terminations}.


\subsection{The Failure-Mode Curriculum}

\begin{figure*}
\vspace{0.25cm}
\begin{subfigure}[b]{0.26\textwidth}
         \centering
         \includegraphics[width=\textwidth]{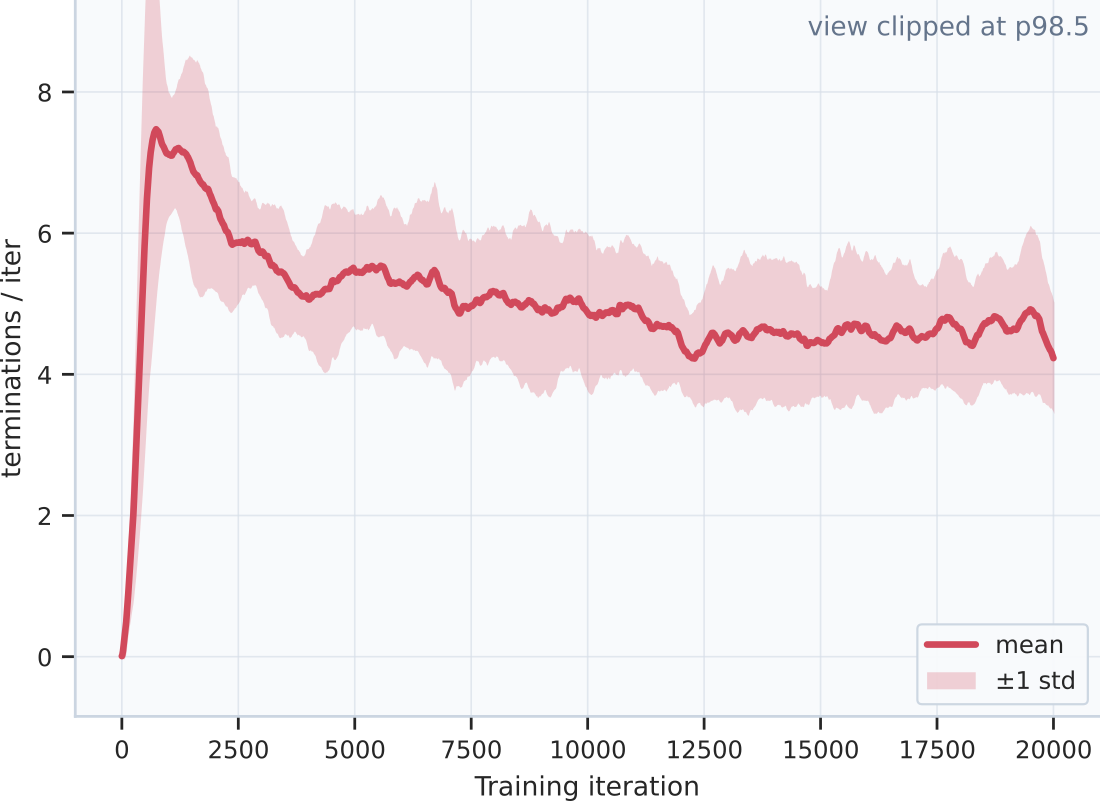}
         \caption{Board OOB}
     \end{subfigure} %
\begin{subfigure}[b]{0.26\textwidth}
         \centering
         \includegraphics[width=\textwidth]{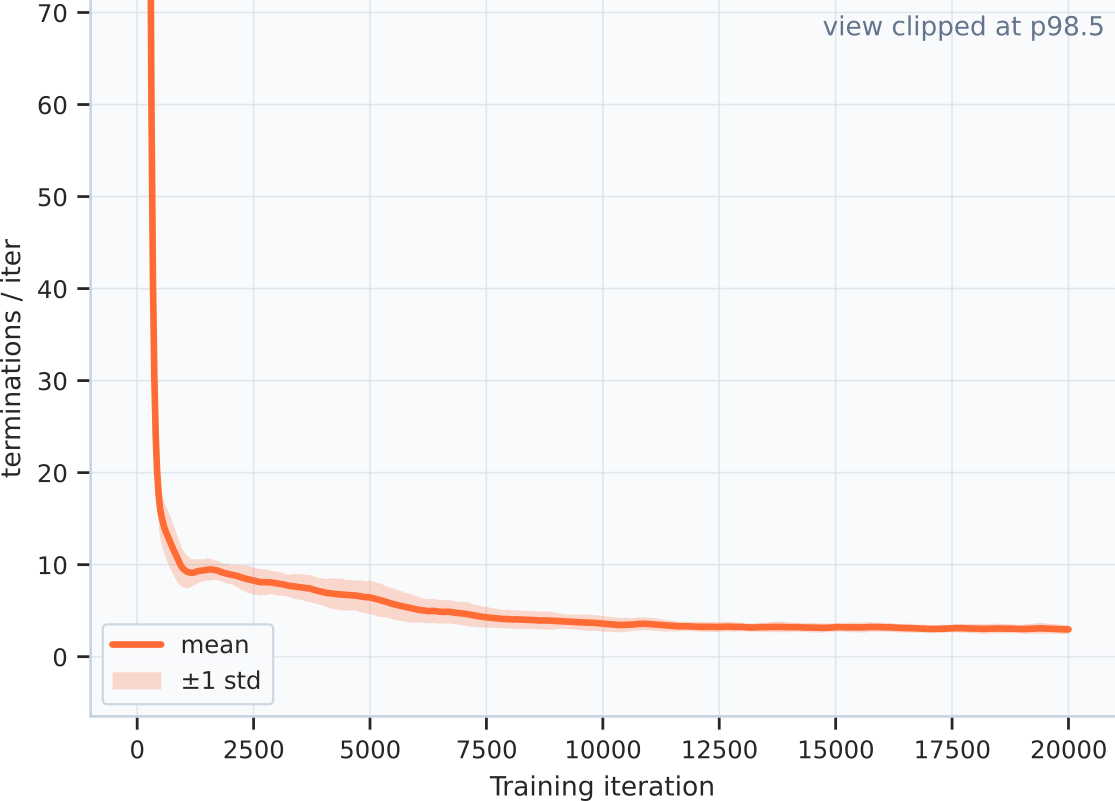}
         \caption{Board foot displaced}
     \end{subfigure} %
\begin{subfigure}[b]{0.26\textwidth}
         \centering
         \includegraphics[width=\textwidth]{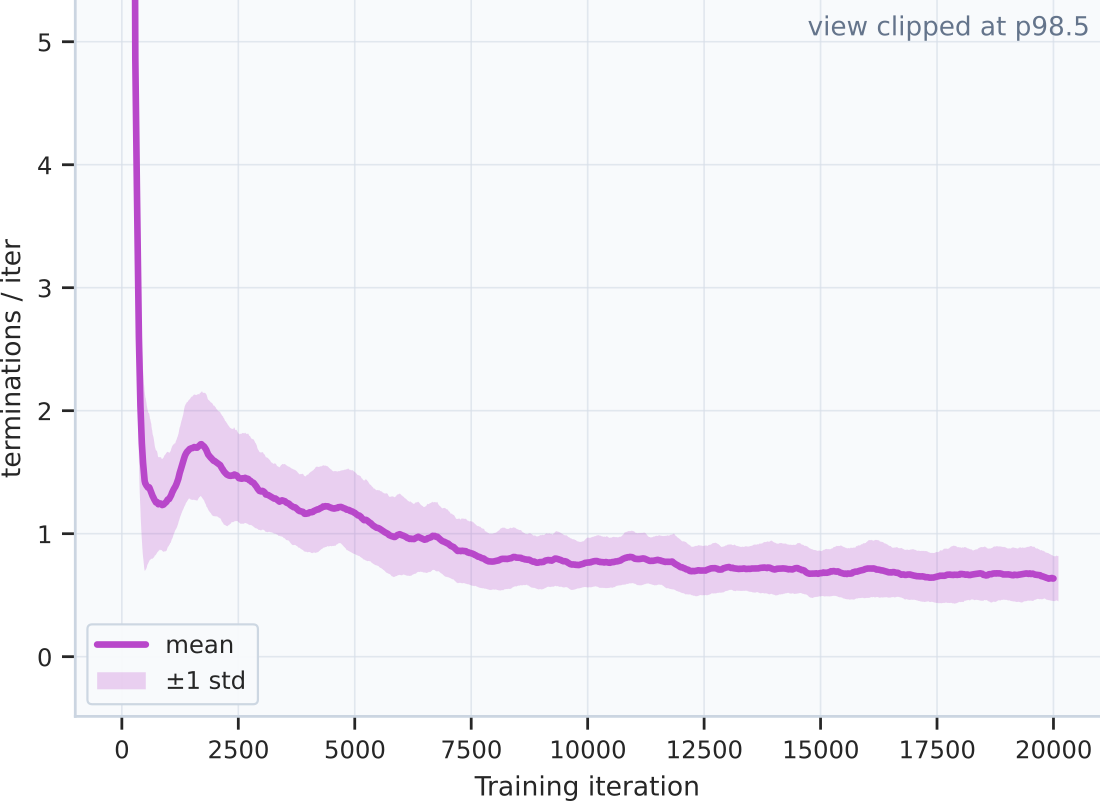}
         \caption{Yaw misaligned}
     \end{subfigure}

\hspace{20ex}
\begin{subfigure}[b]{0.26\textwidth}
         \centering
         \includegraphics[width=\textwidth]{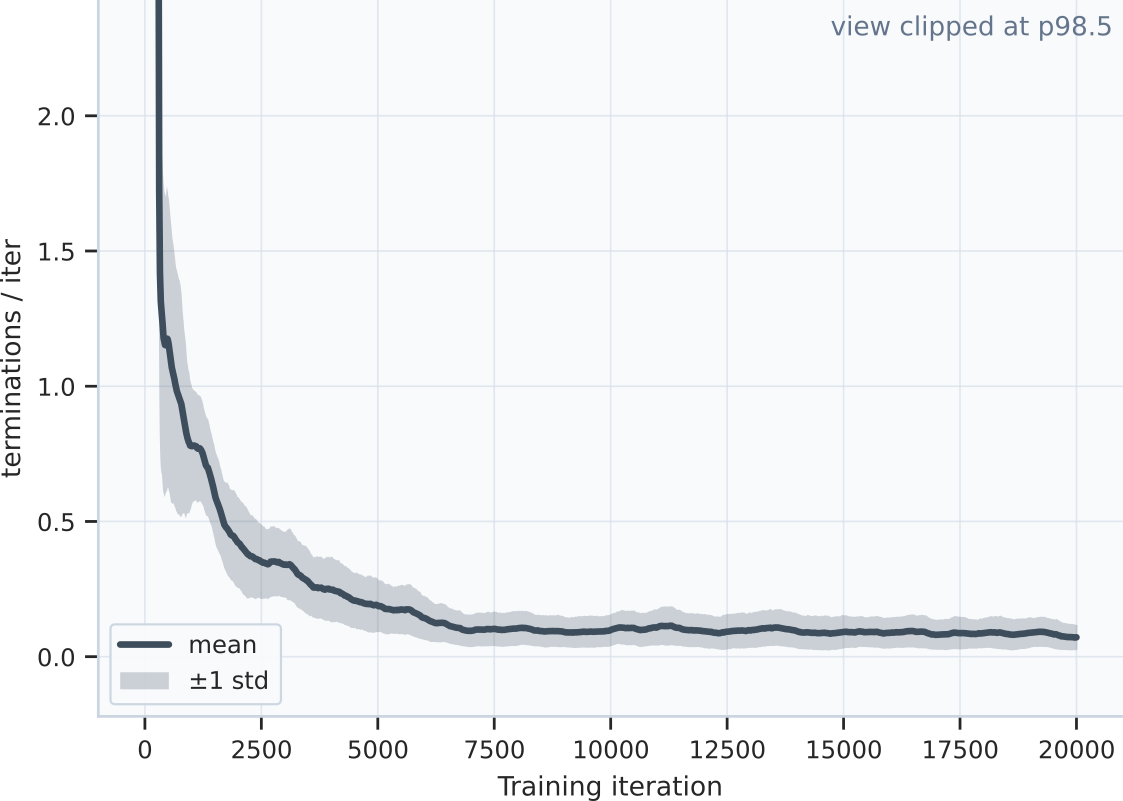}
         \caption{Both feet on board}
     \end{subfigure} %
\begin{subfigure}[b]{0.26\textwidth}
         \centering
         \includegraphics[width=\textwidth]{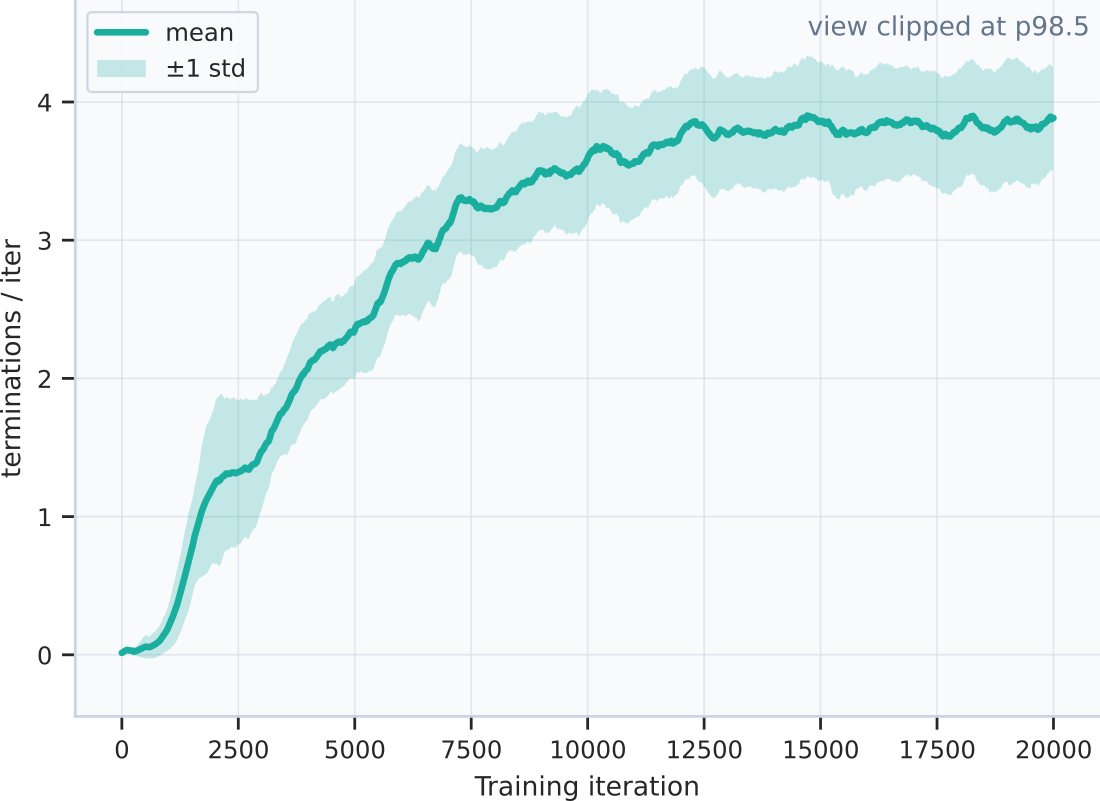}
         \caption{Time-out}
     \end{subfigure} %
\hfill
\caption{\textbf{Failure modes driving skill acquisition.} Per-iteration termination counts (mean $\pm 1$ standard deviation). The dominant failure shifts from board-foot displacement (a stance control problem, early) to board out-of-bounds and time outs (a locomotion problem, late), while both-feet-on-board and yaw exploits stay low-- the implicit curriculum made visible.}
\label{fig:terminations}
\vspace{-0.5cm}
\end{figure*}

Figure~\ref{fig:terminations} shows termination counts over training and supports the proposed implicit curriculum. Initially, board-foot displacement and yaw misalignment dominate because the policy has not yet stabilised its stance. Both decline rapidly, with board-foot displacement continuing to fall throughout training. As the board begins travelling farther, board out-of-bounds rises and becomes the principal locomotion-related failure mode. Time outs also become more frequent, indicating that more episodes survive to the duration limit. Both-feet-on-board and yaw-misalignment terminations remain low after the early phase, suggesting that standing on the deck and excessive sideways twisting become uncommon. Overall, the progression from declining stance failures to coupled motion and longer episodes is consistent with the learning signature proposed in Section~\ref{sec:term}.

\subsection{Performance}

Table~\ref{tab:final} lists the final performance for each experiment. All three achieve high XOR rates and non-zero coupled board and trunk progress, giving a consistent quantitative signature of the learned skill (RQ1, RQ3). Performance nevertheless varies: experiment~0 is fastest and has the most out-of-bounds terminations; experiment~1 is slowest but achieves the longest episodes and highest reward; and experiment~2 is intermediate in speed but has the shortest episodes. 

\begin{table}[t]
\vspace{0.2cm}
\caption{Final scalar outcomes for the three clean seeds.}
\label{tab:final}
\centering
\renewcommand{\arraystretch}{1.05}
\resizebox{\columnwidth}{!}{%
\begin{tabular}{lrrrrrrr}
\toprule
Run & Reward & Ep.\ len. & Board & Push & Trunk & XOR & OOB \\
 & & (steps) & (m/s) & (m/s) & (m/s) & rate & /iter \\
\midrule
0 & 305.96 & 479.96 & 0.359 & 0.394 & 0.405 & 0.981 & 5.83 \\
1 & 352.45 & 582.92 & 0.185 & 0.210 & 0.216 & 0.982 & 2.63 \\
2 & 280.71 & 433.49 & 0.267 & 0.287 & 0.304 & 0.966 & 4.83 \\
\midrule
Mean & 313.04 & 498.79 & 0.270 & 0.297 & 0.308 & 0.976 & 4.43 \\
Std. & 36.39 & 76.47 & 0.087 & 0.092 & 0.095 & 0.009 & 1.64 \\
\bottomrule
\end{tabular}
}
\vspace{-0.4cm}
\end{table}

\subsection{Qualitative Behaviour}

\begin{figure*}[!t]
\vspace{0.5cm}
\centering
\includegraphics[width=0.8\textwidth]{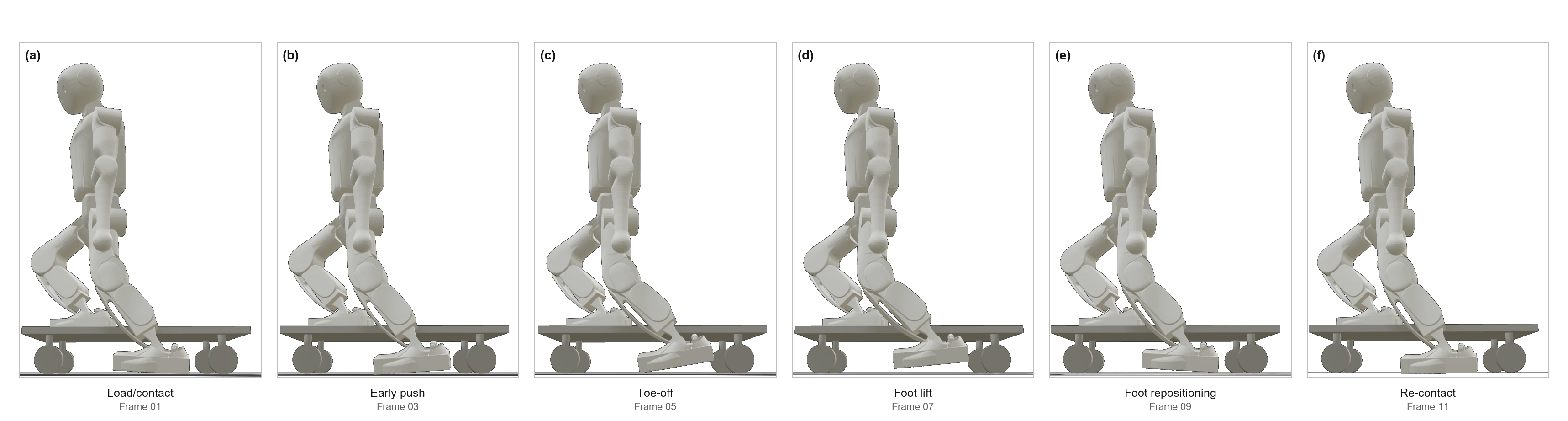}
\caption{\textbf{Side view of the propulsive stepping sequence.} Six frames show the movement of the propulsive foot from load/contact through early push, toe-off, foot lift and repositioning to re-contact.}
\label{fig:push-cycle}
\end{figure*}

\begin{figure*}[!t]
\centering
\includegraphics[width=0.7\textwidth]{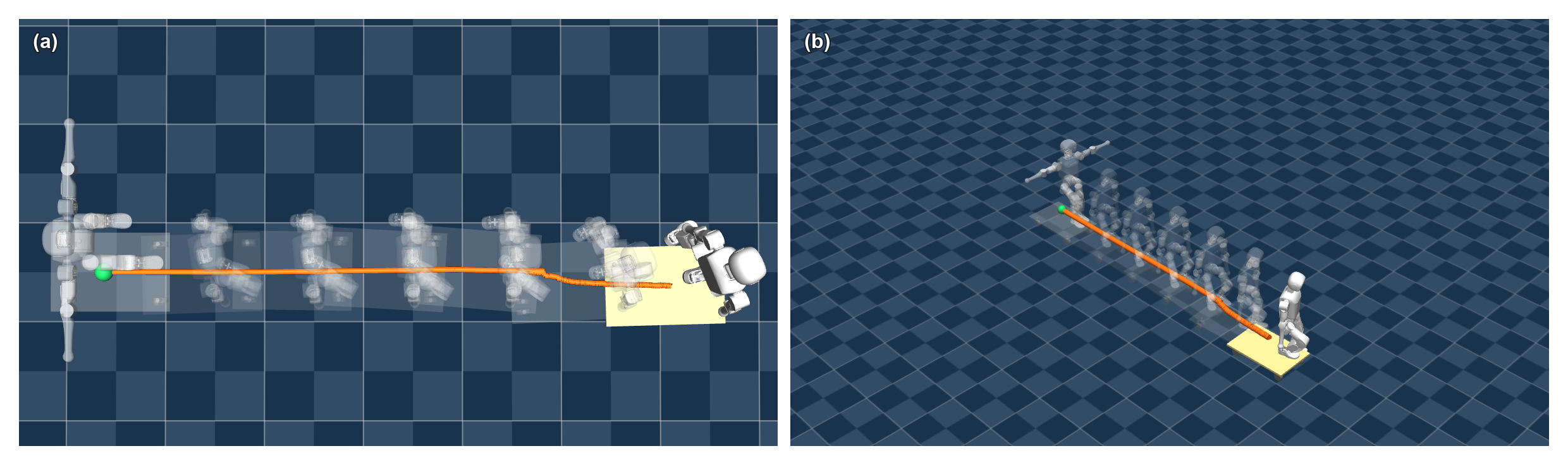}
\caption{\textbf{Motion of the robot--board system through space.} Overhead (left) and oblique-left (right) views show snapshots of the robot and board as they move through the simulation. The orange line traces the board-centre trajectory, and the green marker indicates the starting position.}
\label{fig:spatial-trajectory}
\vspace{-0.2cm}
\end{figure*}

Visual inspection of representative checkpoints supports and qualifies the quantitative results. The policies maintain the push stance, produce forward board motion through ground contact propulsion and remain upright for several hundred steps. Both-feet-on-board and yaw-misalignment terminations become rare, and no previously identified exploit was observed during inspection. However, the motion is conservative: the off-board foot lifts only slightly during each push and remains near the ground for most of the stroke, with a near-ground rate of approximately $0.92$ in representative diagnostic logs. The resulting motion is a small lift-and-replant cycle instead of a fluid, high clearance leg swing. Also, in rare cases, the ground leg rotates inward or pushes with the side of the foot, briefly stalling propulsion. The behaviour is therefore best described as \emph{recognisable but conservative} skateboarding. Section~\ref{sec:why} relates this style to the incentives created by the reward design. Figures~\ref{fig:push-cycle} and~\ref{fig:spatial-trajectory} show the propulsive stepping sequence and the resulting motion through space.

\section{Discussion}
\label{sec:discussion}


Recognisable reference-free skateboard locomotion can be learned in simulation (RQ1), although the learned skill is bounded: all policies keep a near-saturated XOR stance and move the board and trunk forward together through skating-like propulsion rather than board kicking (Section~\ref{sec:results}). The \emph{functional outcome} of skating thus emerges without a skateboarding reference, while its \emph{style} emerges only partially (see below).


Furthermore, the results suggest that the key mechanism is the combination of task rewards and exploit-blocking terminations rather than any single reward term (RQ2). Board-foot displacement appears to promote a planted deck foot, while yaw, both-feet, low-trunk, out-of-bounds, separation and spin terminations constrain non-skating solutions. Within this feasible set, board-distance and grounded-progress rewards favour coupled robot--board motion. The learning pattern predicted in Section~\ref{sec:term} is reflected in Fig.~\ref{fig:terminations}. Thus, the terminations shape which behaviours remain feasible, while the rewards determine which feasible behaviours are preferred.

Finally, the functional skill of skateboarding emerges consistently across all experiments (RQ3), although the speed-stability trade off varies (Table~\ref{tab:final}). Furthermore, the skill can be learned cost-efficiently under computational resource constraints and a time budget of four days.
%
\newline\textbf{Why Conservative Skateboarding Skills Emerge: }
\label{sec:why}
The gap between the learned low-clearance push and a fluid stroke may be partly explained by the reward design. The near-ground term in Eq.~(\ref{eq:nearground}) directly favours low foot clearance. Relative to ground level, its weighted reward rate decreases by approximately $0.24$, $0.88$ and $1.72$ at foot heights of $0.05$, $0.10$, and $0.15$~m, respectively. After multiplication by the $0.02$~s control timestep, these correspond to per-step return reductions of approximately $0.005$, $0.018$ and $0.034$. Although individually small, these reductions accumulate whenever the foot remains elevated. The observed low-clearance motion and near-ground rate of approximately $0.92$ are therefore consistent with this incentive. However, because the term was not evaluated in a controlled ablation, it cannot be identified as the sole cause of the conservative style.

This result also clarifies the contrast with reference-based systems. HUSKY uses an AMP style reward to favour humanlike pushing motions, together with physics guided steering, transition planning and sim-to-real engineering \cite{han2026huskyhumanoidskateboardingphysicsaware}. HUSKY’s more natural motion likely reflects its complete system, including both its motion prior and additional engineering; i.e., it cannot be attributed to AMP alone. The two studies represent different supervision choices: task-based learning without a skateboarding motion prior versus demonstration-guided style learning. 

\section{Limitations and Future Work}
\label{sec:limitations}
The paper marks a first step towards an end-to-end skill learning framework for skateboarding with reinforcement learning. Initial experiments are obtained in simulation and future experiments will consider the investigation of possible sim-to-real challenges of the proposed framework. 
Furthermore, the skateboarding skill has been trained on flat and even terrain, albeit domain randomization has been applied to various physical and control parameters to emulate and compensate for real-world challenges.
The board is also a caster-wheeled transport roller board rather than a skateboard. It has no trucks or kingpin, so leaning the deck does not steer the wheels and the board cannot produce lean-to-steer turning. Our results therefore do not cover truck dynamics, and training on a modelled skateboard, as in HUSKY \cite{han2026huskyhumanoidskateboardingphysicsaware}, is future work.

Future work should first test the causes of the conservative style through controlled reward ablations. Reducing or annealing the near-ground reward, possibly alongside a recovery phase incentive, may encourage a higher-clearance stroke. A lightweight style prior, such as a small AMP discriminator, could instead be explored as a hybrid extension that retains the task rewards and termination structure while adding limited reference-based supervision. Reward and termination ablations, training from scratch and additional seeds would strengthen the causal and statistical evidence, while steering could extend the task beyond straight-line propulsion.

\section{Conclusion}
\label{sec:conclusion}

This paper investigated the extent to which a Booster K1 humanoid can learn to skateboard on a passive roller board without a skateboarding motion reference or target joint-posture trajectory and without more than a single consumer workstation. The result is meaningful but bounded: functional skateboarding emerges, but natural movement style remains only partially developed. PPO training in MuJoCo/mjlab, guided by task rewards and exploit blocking terminations, produced policies that keep one foot on the board and the other near the ground, propel the board forwards and remain upright for hundreds of steps. Across three seeds, the final XOR-stance rate was approximately $97.6\%$, mean board speed was approximately $0.27\mps$ and trunk motion remained coupled to board motion.

Two lessons stand out. First off, early terminations are a useful design tool for object mediated locomotion, because they constrain non-skating, high-return solutions and help direct learning toward the intended mechanics. Secondly, reference-free learning recovered the functional outcome of skating but not its full style: the off-board foot followed a conservative, low-clearance motion rather than a fluid recovery swing. This behaviour is consistent with the incentive created by the near-ground reward, although a controlled ablation is required to establish causality. Ergo this work demonstrates a non trivial humanoid--object skill on modest hardware while identifying clear directions for higher-clearance strokes, steering, stronger experimental validation and eventual sim-to-real transfer.

\section*{Acknowledgements}
This publication is part of the project ETAPE: Embodiment- and Task-Aware Parameter Embeddings for Robotic Foundation Models (file number NGF.1609.243.075) of the research programme AiNed XS Europe, which is financed by the Dutch Research Council (NWO).
\bibliographystyle{ieeetr}
\bibliography{references}

\end{document}